# Approximating the Partition Function by Deleting and then Correcting for Model Edges

**Arthur Choi** and **Adnan Darwiche**
Computer Science Department
University of California, Los Angeles
Los Angeles, CA 90095
{*aychoi,darwiche*}@cs.ucla.edu

## Abstract

We propose an approach for approximating the partition function which is based on two steps: (1) computing the partition function of a simplified model which is obtained by deleting model edges, and (2) rectifying the result by applying an edge-by-edge correction. The approach leads to an intuitive framework in which one can trade-off the quality of an approximation with the complexity of computing it. It also includes the Bethe free energy approximation as a degenerate case. We develop the approach theoretically in this paper and provide a number of empirical results that reveal its practical utility.

## 1 INTRODUCTION

We presented in prior work an approach to approximate inference which is based on performing exact inference on a simplified model (Choi & Darwiche, 2006a, 2006b). We proposed obtaining the simplified model by deleting enough edges to render its treewidth manageable under the current computational resources. Interestingly enough, the approach subsumes iterative belief propagation (IBP) as a degenerate case, and provides an intuitive framework for capturing a class of Generalized Belief Propagation (GBP) approximations (Choi & Darwiche, 2006a; Yedidia, Freeman, & Weiss, 2005).

We show in this paper that the simplified models can also be used to approximate the partition function if one applies a correction for each deleted edge. We propose two edge-correction schemes, each of which is capable of perfectly correcting the partition function when a single edge has been deleted. The first scheme will have this property only when a particular condition holds in the simplified model, and gives rise to the Bethe free energy approximation when applied to a tree-structured approximation (see Yedidia et al., 2005, for more on the Bethe approximation and its relationship to IBP). The second correction scheme does not require such a condition and is shown empirically to lead to more accurate approximations. Both schemes can be applied to the whole spectrum of simplified models and can therefore be used to trade-off the quality of obtained approximations with the complexity of computing them.

This new edge-correction perspective on approximating the partition function has a number of consequences. First, it provides a new perspective on the Bethe free energy approximation, and may serve as a tool to help identify situations when Bethe approximations may be exact or accurate in practice. Next, it suggests that we do not necessarily need to seek good approximations, but instead seek approximations that are accurately correctable. To this end, we propose a heuristic for finding simplified models that is specific to the task of correction. Finally, it provides the opportunity to improve on edge-deletion approximations (and certain GBP approximations), with only a modest amount of computational effort. In particular, we show empirically how it is possible to correct only for a small number of edges that have the most impact on an approximation.

Proofs of results appear in the Appendix.

## 2 EDGE DELETION

We first review our edge deletion framework in probabilistic graphical models. For simplicity, we consider pairwise Markov random fields, although our framework can easily be extended to general Markov networks as well as to factor graphs. For an application to directed models, see (Choi & Darwiche, 2006a).

Let a pairwise Markov random field (MRF) $\mathcal{M}$ have a graph $(\mathcal{E}, \mathcal{V})$ with edges $(i,j) \in \mathcal{E}$ and nodes $i \in \mathcal{V}$, where each node $i$ of the graph is associated with a variable $X_i$ taking on values $x_i$. Edges $(i,j)$ are associated with edge potentials $\psi(x_i, x_j)$ and nodes $i$ with node potentials $\psi(x_i)$. The (strictly positive) distribution $Pr$ induced by $\mathcal{M}$ is defined as follows:

$$Pr(\mathbf{x}) \stackrel{def}{=} \frac{1}{Z} \prod_{(i,j) \in \mathcal{E}} \psi(x_i, x_j) \prod_{i \in \mathcal{V}} \psi(x_i),$$

where $\mathbf{x}$ is an instantiation $x_1, \ldots, x_n$ of network variables, and where $Z$ is the *partition function*:

$$Z \stackrel{def}{=} \sum_{\mathbf{x}} \prod_{(i,j) \in \mathcal{E}} \psi(x_i, x_j) \prod_{i \in \mathcal{V}} \psi(x_i).$$

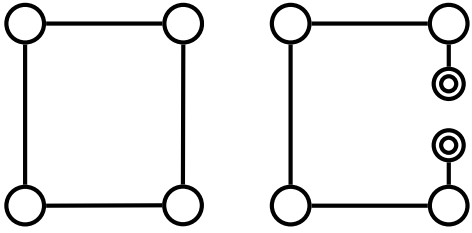

Figure 1: An MRF (left); after edge deletion (right).

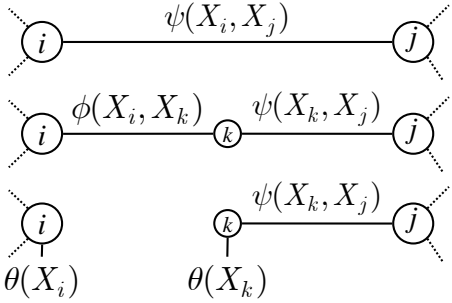

Figure 2: To delete edge $(i,j)$ (top), we introduce auxiliary node $k$ (middle), and delete equivalence edge $(i,k)$, adding edge parameters (bottom).

The basic idea behind our framework is to delete enough edges from the pairwise MRF to render it tractable for exact inference.

**Definition 1** *Let $\mathcal{M}$ be a pairwise MRF. To delete edge $(i,j)$ from $\mathcal{M}$ we remove the edge $(i,j)$ from $\mathcal{M}$ and then introduce the auxiliary potentials $\theta(X_i)$ and $\theta(X_j)$ for variables $X_i$ and $X_j$.*

Figure 1 provides an example of deleting an edge. When deleting multiple edges, note that we may introduce multiple, yet distinct, potentials $\theta(X_i)$ for the same node $X_i$. We shall refer to auxiliary potentials $\theta(X_i)$ and $\theta(X_j)$ as *edge parameters* and use $\Theta$ to denote the set of all edges parameters. The resulting pairwise MRF will be denoted by $\mathcal{M}'(\Theta)$, its partition function will be denoted by $Z'(\Theta)$ and its distribution will be denoted by $Pr'(.; \Theta)$. When choosing a particular value for edge parameters $\Theta$, we will drop reference to $\Theta$, using only $\mathcal{M}'$, $Z'$ and $Pr'(.)$.

Note that while the distribution $Pr(.)$ and partition function $Z$ of the original pairwise MRF $\mathcal{M}$ may be hard to compute, the distribution $Pr'(.; \Theta)$ and partition function $Z'(\Theta)$ of $\mathcal{M}'(\Theta)$ should be easily computable due to edge deletion. Note also that before we can use $Pr'(.; \Theta)$ and $Z'(\Theta)$ to approximate $Pr(.)$ and $Z$, we must first specify the edge parameters $\Theta$. In fact, it is the values of these parameters which will control the quality of approximations $Pr'(.; \Theta)$ and $Z'(\Theta)$.

Without loss of generality, we will assume that we are only deleting *equivalence edges* $(i,j)$, which connect two variables $X_i$ and $X_j$ with the same domain, and have a potential $\phi(x_i, x_j)$ that denotes an equivalence constraint: $\phi(x_i, x_j) = 1$ if $x_i = x_j$, and $\phi(x_i, x_j) = 0$ otherwise. The deletion of any edge in an MRF can be formulated as the deletion of an equivalence edge.[1]

As for the values of the edge parameters, we proposed (and justified) in (Choi & Darwiche, 2006a) the following conditions on $\theta(x_i)$ and $\theta(x_j)$:

$$\theta(x_i) = \alpha \frac{\partial Z'}{\partial \theta(x_j)} \quad \text{and} \quad \theta(x_j) = \alpha \frac{\partial Z'}{\partial \theta(x_i)} \qquad (1)$$

where $\alpha$ is a normalizing constant. Note that the partial derivatives of Equation 1 can be computed efficiently in traditional inference frameworks (Darwiche, 2003; Park & Darwiche, 2004).

Equation 1 can also be viewed as update equations, suggesting an iterative method that searches for edge parameters, which we called ED-BP (Choi & Darwiche, 2006a). Starting with an initial approximation $\mathcal{M}'_0$ at iteration $t = 0$ (say, with uniform parameters), we can compute edge parameters $\theta_t(x_i)$ and $\theta_t(x_j)$ for an iteration $t > 0$ by performing exact inference in the approximate network $\mathcal{M}'_{t-1}$. We repeat this process until we observe that all parameters converge to a fixed point satisfying Equation 1 (if ever).

Note that Equation 1 does not specify a unique value of edge parameters, due to the constants $\alpha$. That is, each value of these constants will lead to a different set of edge parameters. Yet, independent of which constants we use, the resulting pairwise MRF $\mathcal{M}'$ will have an *invariant* distribution $Pr'(.)$ that satisfies the following properties. First,

$$Pr'(x_i) = Pr'(x_j) = \frac{1}{z_{ij}} \cdot \theta(x_i)\theta(x_j), \qquad (2)$$

where $z_{ij} = \sum_{x_i=x_j} \theta(x_i)\theta(x_j)$. Next, if the pairwise MRF $\mathcal{M}'$ has a tree structure, the node and edge marginals of distribution $Pr'(.)$ will correspond precisely to the marginals obtained by running IBP on the original model $\mathcal{M}$. Moreover, if the pairwise MRF $\mathcal{M}'$ has loops, the node marginals of distribution $Pr'$ will correspond to node marginals obtained by running generalized belief propagation (GBP) using a particular joingraph for the original model $\mathcal{M}$ (Yedidia et al., 2005; Choi & Darwiche, 2006a).

---

[1] To delete an MRF edge $(i,j)$ that is not an equivalence edge, we use the technique illustrated in Figure 2: we introduce an auxiliary node $k$ between $i$ and $j$; introduce an equivalence constraint on the edge $(i,k)$; copy the original potential of edge $(i,j)$ to $(k,j)$; and delete the equivalence edge $(i,k)$. Note that the original model and the extended one will: (1) have the same treewidth, (2) agree on the distribution over their common variables, and (3) have the same partition function values.

## 3 EDGE CORRECTION

While the edge parameters specified by Equation 1 are guaranteed to yield an invariant distribution $Pr'(.)$, they are not guaranteed to yield an invariant partition function $Z'$ as this function is sensitive to the choice of constants $\alpha$. Hence, while these edge parameters will yield an interesting approximation of node marginals, they do not yield a meaningful approximation of the partition function.

We will show in this section, however, that one can apply an edge-by-edge correction to the partition function $Z'$, leading to a corrected partition function that is invariant to the choice of constants $\alpha$. This seemingly subtle approach leads to two important consequences. First, it results in a semantics for the Bethe free energy approximation as a corrected partition function. Second, it allows for an improved class of approximations based on improved corrections.

### 3.1 ZERO EDGE-CORRECTION

We will now propose a correction to the partition function $Z'$, which gives rise to the Bethe free energy and some of its generalizations.

**Proposition 1** *Let $\mathcal{M}'$ be the result of deleting a single equivalence edge $(i,j)$ from a pairwise MRF $\mathcal{M}$. If the parameters of edge $(i,j)$ satisfy Equation 1, and if the mutual information between $X_i$ and $X_j$ in $\mathcal{M}'$ is zero, then:*

$$Z = Z' \cdot \frac{1}{z_{ij}}, \quad \text{where} \quad z_{ij} = \sum_{x_i=x_j} \theta(x_i)\theta(x_j).$$

That is, if we delete a *single* edge $(i,j)$ and find that $X_i$ and $X_j$ are independent in the resulting model $\mathcal{M}'$, we can correct the partition function $Z'$ by $z_{ij}$ and recover the exact partition function $Z$. Moreover, the result of this correction is invariant to the constants $\alpha$ used in Equation 1.

From now on, we will use $MI(X_i;X_j)$ to denote the mutual information between two variables $X_i$ and $X_j$, computed in the *simplified* MRF $\mathcal{M}'$. Moreover, when $MI(X_i;X_j) = 0$, we will say that the deleted edge $(i,j)$ is a zero-MI edge. Note that while an edge may be zero-MI in $\mathcal{M}'$, the mutual information between $X_i$ and $X_j$ in the original MRF $\mathcal{M}$ may still be high.

Let us now consider the more realistic situation where we delete multiple edges, say $\mathcal{E}^\star$, from $\mathcal{M}$ to yield the model $\mathcal{M}'$. We propose to accumulate the above correction for each of the deleted edges, leading to a corrected partition function $Z' \cdot \frac{1}{z}$, where

$$z = \prod_{(i,j)\in\mathcal{E}^\star} z_{ij} = \prod_{(i,j)\in\mathcal{E}^\star} \sum_{x_i=x_j} \theta(x_i)\theta(x_j). \qquad (3)$$

We will refer to this correction as a *zero-MI edge correction*, or EC-Z. This correction is no longer guaranteed to recover the exact partition function $Z$, even if each of the deleted edges is a zero-MI edge. Yet, if the pairwise MRF $\mathcal{M}'$ has a tree structure, applying this correction to the partition function $Z'$ gives rise to the Bethe free energy approximation.

To review, the Bethe free energy $F_\beta$ is an approximation of the true free energy $F$ of a pairwise MRF $\mathcal{M}$, and is exact when $\mathcal{M}$ has a tree structure (Yedidia et al., 2005). In this case, $F = -\log Z$, so we can in principle use $F_\beta$ as an approximation of the partition function $Z$, even when $\mathcal{M}$ does not have a tree structure, i.e., we can use $Z_\beta = \exp\{-F_\beta\}$.

**Theorem 1** *Let $\mathcal{M}'$ be the result of deleting equivalence edges from a pairwise MRF $\mathcal{M}$. If $\mathcal{M}'$ has a tree structure and its edge parameters are as given by Equation 1, we have $Z_\beta = Z' \cdot \frac{1}{z}$.*

Hence, the Bethe approximation of $Z$ is a degenerate case of the EC-Z correction. Thus, IBP and the closely related Bethe approximation, which are exact when an MRF $\mathcal{M}$ is a tree, are naturally characterized by tree-structured ED-BP approximations $\mathcal{M}'$. In particular, exact inference in the simplified network $\mathcal{M}'$ yields: (1) node and edge marginals that are precisely the approximate marginals given by IBP (Choi & Darwiche, 2006a), and now (2) a rectified partition function that is precisely the Bethe approximation; cf. (Wainwright, Jaakkola, & Willsky, 2003).[2]

---

[2]Wainwright et al. proposed tree-based reparametrization (TRP), an algorithm that iteratively reparameterizes the node and edge potentials of a pairwise MRF. At convergence, the node and edge potentials of a tree (any tree)

Since the EC-Z correction is specified purely in quantities available in the model $\mathcal{M}'$, it will be easily computable as long as the model $\mathcal{M}'$ is sparse enough (i.e., it has a treewidth that is manageable under the given computational resources). Hence, this correction can be practically applicable even if $\mathcal{M}'$ does not have a tree structure. In such a case, the correction will lead to an approximation of the partition function which is superior to the one obtained by the Bethe free energy. We will illustrate this point empirically in Section 6.

### 3.2 GENERAL EDGE-CORRECTION

Proposition 1 gives us a condition that allows us to correct the partition function exactly, but under the assumption that the single edge deleted is zero-MI. The following result allows us, in fact, to correct the partition function when deleting *any* single edge.

**Proposition 2** *Let $\mathcal{M}'$ be the result of deleting a single equivalence edge $(i,j)$ from a pairwise MRF $\mathcal{M}$. If the parameters of edge $(i,j)$ satisfy Equation 1, then:*

$$Z = Z' \cdot \frac{y_{ij}}{z_{ij}}, \quad \text{where} \quad y_{ij} = \sum_{x_i = x_j} Pr'(x_i \mid x_j).$$

Note that when the deleted edge $(i,j)$ happens to be zero-MI, factor $y_{ij}$ is 1, and thus Proposition 2 reduces to Proposition 1.

We can also use this proposition as a basis for correcting the partition function when multiple edges are deleted, just as we did in Equation 3. In particular, we now propose using the correction $Z' \cdot \frac{y}{z}$, where $z$ is the same factor given in Equation 3, and

$$y = \prod_{(i,j) \in \mathcal{E}^\star} y_{ij} = \prod_{(i,j) \in \mathcal{E}^\star} \sum_{x_i = x_j} Pr'(x_i \mid x_j), \quad (4)$$

which we refer to as a *general edge correction*, or EC-G.

We note that when every edge is deleted in an ED-BP network $\mathcal{M}'$, every deleted edge becomes a zero-MI edge. Thus, in this case, EC-G reduces to EC-Z, and both yield the Bethe free energy, as in Theorem 1. As we recover more edges, we may expect EC-G to offer

---

embedded in the reparametrized MRF induces a distribution whose exact node and edge marginals are consistent with the corresponding marginals given by IBP. In contrast to ED-BP, TRP's embedded-tree distributions are already normalized, i.e., their partition function is 1. Moreover, generalizations of TRP appeal to auxiliary representations, via reparametrization in joingraphs and hypergraphs. In contrast, the semantics of ED-BP suggest that we simply delete fewer edges. As we shall see in Section 5, the semantics of edge correction further suggest intuitive edge recovery heuristics for choosing more structured approximations.

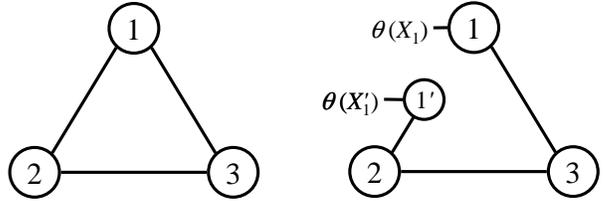

Figure 3: An MRF (left); after deleting edge $(1,2)$, as in Figure 2 (right).

improved approximations over EC-Z, as it relaxes the zero-MI assumption for deleted edges. Accordingly, we may want to delete different edges for EC-G than we would for EC-Z.

## 4 AN EXAMPLE

We provide here an illustrative example of our edge correction techniques. Consider a network of three nodes $X_1, X_2$ and $X_3$ that form a clique, with the following edge potentials:

| $X_i$ | $X_j$ | $\psi(X_1, X_2)$ | $\psi(X_1, X_3)$ | $\psi(X_2, X_3)$ |
|---|---|---|---|---|
| $x_i$ | $x_j$ | .9 | .1 | .081 |
| $x_i$ | $\bar{x}_j$ | .1 | .9 | .810 |
| $\bar{x}_i$ | $x_j$ | .1 | .9 | .090 |
| $\bar{x}_i$ | $\bar{x}_j$ | .9 | .1 | .900 |

Suppose now that we delete the edge $(1,2)$ by replacing $(1,2)$ with a chain $\{(1,1'),(1',2)\}$ and deleting the equivalence edge $(1,1')$; see Figure 3. Using ED-BP to parameterize this deleted edge, we have (to 4 digits):

| $X_i$ | $\theta(X_1)$ | $\theta(X_1')$ |
|---|---|---|
| $x_i$ | .4789 | .8273 |
| $\bar{x}_i$ | .5211 | .1727 |

and we compute $Z' \approx 0.4447$. In this example, edge $(1,1')$ happens to be a zero-MI edge, so $y_{ij} = 1$ and $z_{ij} \approx 0.4862$. Further, we know that both Propositions 1 and 2 allow us to recover the true partition function $Z = Z' \cdot \frac{1}{z_{ij}} \approx 0.9146$.

Now, suppose that we replace the potential on edge $(2,3)$ with $1 - \psi(X_2, X_3)$. In this case, ED-BP gives us edge parameters (to 4 digits):

| $X_i$ | $\theta(X_1)$ | $\theta(X_1')$ |
|---|---|---|
| $x_i$ | .5196 | .1951 |
| $\bar{x}_i$ | .4804 | .8049 |

and we compute $Z' \approx 0.5053$. In this case, edge $(1,1')$ is not a zero-MI edge. Here, we find that $y_{ij} \approx 1.0484$ and $z_{ij} \approx 0.4880$. Since we only delete a single edge, Proposition 2 recovers the true partition function $Z = Z' \cdot \frac{y_{ij}}{z_{ij}} = 1.08542$ whereas Proposition 1 gives only an approximation $Z' \cdot \frac{1}{z_{ij}} \approx 1.0353$.

## 5 EDGE RECOVERY

Suppose we already have a tree-structured approximation $\mathcal{M}'$ of the original model $\mathcal{M}$, but are afforded more computational resources. We can then improve the approximation by *recovering* some of the deleted edges. However, which edge's recovery would have the most impact on the quality of the approximation?

**Edge Recovery for EC-Z.** Since EC-Z is exact for a single deleted edge when $MI(X_i; X_j) = 0$, one may want to recover those edges $(i, j)$ with the highest mutual information $MI(X_i; X_j)$. In fact, this is the same heuristic proposed by (Choi & Darwiche, 2006a) for improving marginal approximations. We will indeed show the promise of this heuristic for EC-Z corrections, in Section 6. On the other hand, we also show that it turns out to be a poor heuristic for EC-G corrections.

**Edge Recovery for EC-G.** Consider the situation when two equivalence edges are deleted, $(i, j)$ and $(s, t)$. In this case, we use the approximate correction:

$$Z' \cdot \frac{y}{z} = Z' \cdot \frac{y_{ij}}{z_{ij}} \frac{y_{st}}{z_{st}},$$

where $\frac{y_{ij}}{z_{ij}}$ is the single-edge correction for edge $(i, j)$ and $\frac{y_{st}}{z_{st}}$ is the single-edge correction for edge $(s, t)$.

The question now is: When is this double-edge correction exact? Intuitively, we want to identify a situation where each edge can be corrected, independently of the other. Consider then the case where variables $X_i, X_j$ are independent of variables $X_s, X_t$ in $\mathcal{M}'$.

**Proposition 3** *Let $\mathcal{M}'$ be the result of deleting two equivalence edges, $(i, j)$ and $(s, t)$, from a pairwise MRF $\mathcal{M}$. If the edge parameters of $\mathcal{M}'$ satisfy Equation 1, and if $MI(X_iX_j; X_sX_t) = 0$ in $\mathcal{M}'$, then:*

$$Z = Z' \cdot \frac{y_{ij}}{z_{ij}} \frac{y_{st}}{z_{st}}.$$

This suggests a new edge recovery heuristic for EC-G approximations to the partition function. Initially, we start with a tree-structured network $\mathcal{M}'$. We assign each deleted edge $(i, j)$ a score:

$$\sum_{(s,t) \in \mathcal{E}^\star \setminus (i,j)} MI(X_iX_j; X_sX_t).$$

We then prefer to recover the top $k$ edges with the highest mutual information scores.

## 6 EXPERIMENTS

Our goal here is to highlight different aspects of edge-correction, edge-recovery, and further a notion of partial correction. Starting from a random spanning tree (dropping instances where ED-BP and hence IBP, do not converge), we rank each deleted edge, and recover edges $k$ at a time until all edges are recovered. At each point, we evaluate the quality of the approximation by the average relative error $|\widehat{Z} - Z|/Z$, where $\widehat{Z}$ denotes the designated approximation. Remember that in a tree-structured approximation, when no edge is recovered, EC-Z corresponds to the Bethe approximation. Likewise, when every edge is recovered, both EC-Z and EC-G are exact. Although, for simplicity, we presented our edge-correction framework in the context of pairwise MRFs, some of our experiments are run on Bayesian networks, to which all of our results also apply.[3] In these cases, observations are generated from the joint distribution over all leaves, unless otherwise specified.

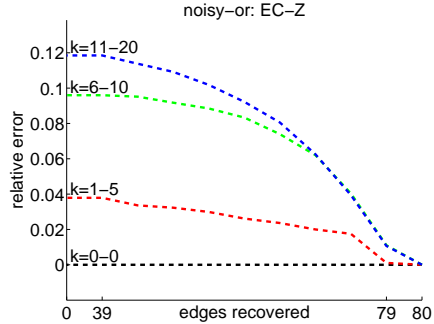

Figure 4: Edge correction in noisy-or networks.

**Noisy-or.** We consider first random two-layer noisy-or networks. Deleting an edge in this network effectively disconnects a cause variable $C$ from an effect variable $E$, where a clone $\hat{C}$ replaces $C$ as a a cause of $E$.[4] In this situation, we may use edge-correction to reason how well EC-Z and the Bethe approximation may perform. With no positive findings, for example, we know that all causes are pairwise mutually independent, including a cause $C$ and its clone $\hat{C}$ in a noisy-or network where edges have been deleted. Starting from a tree-structured approximation, corresponding to the Bethe approximation, every recoverable edge is zero-MI and will remain zero-MI up to the point where all edges are recovered. Thus we may infer EC-Z to be exact throughout, and thus also that the Bethe approximation is exact.

Consider now Figure 4, which compares the quality of EC-Z corrections as edges are recovered randomly. We generated over 400 random noisy-or networks,[5] where

---
[3]Most of the Bayesian networks used here are available at http://www.cs.huji.ac.il/labs/compbio/Repository.
[4]As in Section 2, we replace edge $C \to E$ with a chain $C \to \hat{C} \to E$, and delete the equivalence edge $C \to \hat{C}$.
[5]Each network was given 20 roots and 20 sinks, where

for each network, we randomly chose $k$ of 20 effect variables as positive findings and the remaining effect variables as negative findings. We have 4 cases here measuring the quality of the EC-Z approximation, each an average over a range of positive findings: 0, 1–5, 6–10, 11–20. As predicted, the EC-Z and Bethe approximations are exact with 0 positive findings. Given this, we expect, and observe, that with more positive findings, and fewer zero-MI edges, the EC-Z and Bethe approximations tend to be less accurate.

**Edge recovery.** Consider now Figure 5, where we compare EC-Z corrections to EC-G corrections, but also the impact that different edge recovery heuristics can have on an approximation. Here, plots are averages of over 50 instances. In the first plot, we took random $6 \times 6$ grid networks, where pairwise couplings were randomly given parameters in $[0.0, 0.1)$ or $(0.9, 1.0]$. First, when we compare EC-Z and EC-G by random edge recovery, we see that EC-G is a notable improvement over EC-Z, even when no edges are recovered. When we use the mutual information heuristic (MI) designed for EC-Z, the EC-Z approximations also improve considerably. However, EC-G approximations are worse than when we randomly recovered edges! Although EC-G approximations still dominate the EC-Z ones, this example illustrates that EC-Z approximations (based on the Bethe approximation) and EC-G approximations (based on exact corrections for a single edge) are of a different nature, and suggest that an alternative approach to recovery may be needed. Indeed, when we use the mutual information heuristic (MI2) designed for EC-G, we find that EC-G easily dominates the first four approximations. We see similar results in the `win95pts` and `water` networks.

**Partial corrections.** Although the individual edge-corrections for EC-Z are trivial to compute, the corrections for EC-G require joint marginals. In the case where we need to correct for many deleted edges, the EC-G corrections of Equation 4 may become expensive to compute. We may then ask: Can we effectively improve an approximation, by correcting for only a subset of the edges?

Consider then Figure 6, where we plot how the quality of our approximation evolves over time (averaged over 50 instances), over two steps: (1) the ED-BP parametrization algorithm, and after convergence (2) EC-G edge correction. On the first half of each plot, we start with a tree-structured approximate network, and compute the EC-Z approximation as ED-BP (and equivalently, IBP, in this case) runs for a fixed number of iterations. Eventually, the edge-corrected partition function converges (to the Bethe approximation), at

---

sinks are given 4 random parents. Network parameters were also chosen randomly.

which point we want to compute the edge corrections for EC-G. We can compute the corrections for an edge, one-by-one, applying them to the EC-Z approximation as they are computed. Since edge corrections are invariant to the order in which they are computed, we can then examine a notion of a *partial* EC-G approximation that accumulates only the correction factors for a given subset of deleted edges.

On the right half of each plot, we compute the error in a partial EC-G approximation given two separate orderings of deleted edges. The first ordering, which we consider to be "optimal", pre-computes corrections for all edges and sorts them from largest to smallest. In the `win95pts` network, we find that in fact, most of the edges have very little impact on the final EC-G approximation! Moreover, the time it took to compute the most important corrections required only as much time as it took ED-BP (IBP) to converge. This suggests that it is possible to improve on the Bethe approximation, with only a modest amount of additional computation (in the time to compute corrections for the important edges).

Of course, such an approach would require a way to identify the most important corrections, without actually computing them. In (Choi & Darwiche, 2008), we proposed a soft extension of d-separation in polytree Bayesian networks that was shown to be effective in ranking edges for the process of edge recovery (as in EC-Z). Applying it here to the task of ranking edge-corrections, we find that it is also effective at identifying important edges for correction. For example, in the `win95pts` network, soft d-separation (sd-sep) is nearly as competitive with the "optimal" at producing partial EC-G approximations. Moreover, soft d-separation is much more efficient, requiring only node and edge marginals to rank *all* deleted edges.

We see a similar story in the `pigs` and `mildew` network. In the `mildew` network, where many deleted edges have an impact on the approximation, the quality of the approximation tends to improve monotonically (on average), so we may still desire to perform as many individual corrections as resources allow.

## 7  EDGE CORRECTIONS AND FREE ENERGIES

As the Bethe free energy is an edge-corrected partition function, EC-Z and EC-G approximations can be viewed also from the perspective of free energies.

When the model $\mathcal{M}'$ is a tree, EC-Z yields the influential Bethe free energy approximation (Yedidia et al., 2005). When the model $\mathcal{M}'$ has cycles, it can be shown that EC-Z corresponds more generally to jo-ingraph free energy approximations (Aji & McEliece, 2001; Dechter, Kask, & Mateescu, 2002); see (Choi & Darwiche, 2006a) for the connection to iterative join-graph propagation.

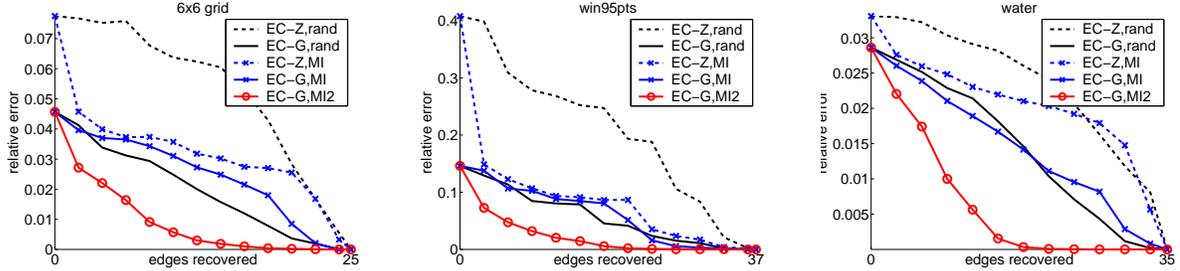

Figure 5: EC-Z versus EC-G, and edge recovery.

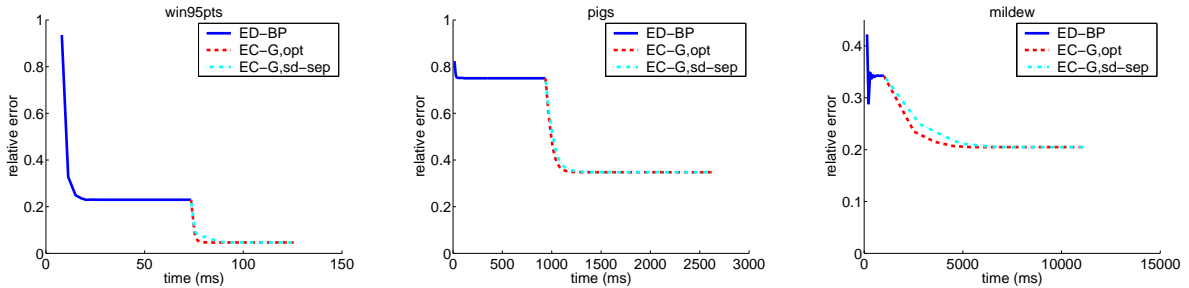

Figure 6: Time to parametrize by ED-BP, and compute EC-G corrections.

The EC-G correction can also take the form of another free energy approximation. Note first that when multiple equivalence edges are deleted, we can compute the partition function $Z'_{ij}$ of a model $\mathcal{M}'_{ij}$ where the single edge $(i,j)$ has been recovered (keeping edge parameters for all other edges fixed): $Z'_{ij} = Z' \cdot \frac{y_{ij}}{z_{ij}}$. Therefore, we have that:

$$Z' \cdot \frac{y}{z} = Z' \cdot \prod_{(i,j) \in \mathcal{E}^\star} \frac{y_{ij}}{z_{ij}} = Z' \cdot \prod_{(i,j) \in \mathcal{E}^\star} \frac{Z'_{ij}}{Z'}.$$

This yields a (dual) energy of the form $-\log(Z' \cdot \frac{y}{z}) = (n-1)\log Z' - \sum_{(i,j) \in \mathcal{E}^\star} \log Z'_{ij}$, where $n$ is the number of equivalence edges $(i,j)$ deleted. Whereas we fixed, somewhat arbitrarily, our edge parameters to satisfy Equation 1, we could in principle seek edge parameters optimizing the above free energy directly, giving rise to EP and GBP free energy approximations with higher-order structure (Welling, Minka, & Teh, 2005). On the other hand, edge recovery heuristics for EC-G could possibly serve as a heuristic for identifying improved EP and GBP free energies, directly. This is a perspective that is currently being investigated.

While we are concerned mostly with IBP and the closely related Bethe free energy approximation, we expect that an edge-correction perspective may be useful in improving other reasoning algorithms, particularly those that can be formulated as exact inference in simplified models. These include, as we have shown here, IBP and some of its generalizations (Yedidia et al., 2005), but also numerous variational methods (Jordan, Ghahramani, Jaakkola, & Saul, 1999; Wiegerinck, 2000; Geiger, Meek, & Wexler, 2006) and their corresponding free energy approximations. Also related, is tree-reweighted belief propagation (TRW) (Wainwright, Jaakkola, & Willsky, 2005), which provides upper bounds on the log partition function, and can be thought of as a convexified form of the Bethe free energy. Mean field methods and its generalizations are another well-known class of approximations that provide lower bounds on the partition function (e.g., Saul & Jordan, 1995; Jaakkola, 2001). Although the latter have been found to be useful, others have found that the Bethe free energy can often provide better quality approximations, (e.g., Weiss, 2001). Similarly, comparing EC-Z approximations and mean-field bounds derived from approximations with the same structure, we find that EC-Z, which does not guarantee bounds, offers better approximations.

## 8 CONCLUSION

We proposed an approach for approximating the partition function which is based on two steps: (1) computing the partition function of a simplified model which is obtained by deleting model edges, and (2) rectifying the result by applying an edge-by-edge correction. The approach leads to an intuitive framework in which one can trade-off the quality of an approximation with the complexity of computing it through a simple process of edge recovery. We provided two concrete instantiations of the proposed framework by proposing two edge correction schemes with corresponding heuristics for edge recovery. The first of these instantiations captures the well known Bethe free energy approximation as a degenerate case. The second instantiation has been shown to lead to more accurate approximations, more so when edge recovery is targeted towards accurate correction. We further highlighted, in our experiments, how edge correction could be used as a conceptual tool to help identify situations where the Bethe approximation may be exact, or accurate. Finally, we suggested a notion of partial correction, that can improve on the Bethe approximation with only a modest amount of computational effort.

**Acknowledgments**

This work has been partially supported by Air Force grant #FA9550-05-1-0075 and by NSF grant #IIS-0713166.

## A  PROOFS

Note that Proposition 1 follows from Proposition 2.

**Proof of Theorem 1** When a given model is a tree, the Bethe free energy is exact. We then consider the exact energy of a tree-structured $\mathcal{M}'$ where $F' = -\log Z'$. Our goal then is to show that $Z_\beta = Z' \cdot \frac{1}{z}$, or equivalently, $F' = F_\beta - \log z$.

Let $E[\,.\,]$ denote expectations and $H(.)$ denote entropies with respect to IBP beliefs, and equivalently, ED-BP marginals in $\mathcal{M}'$ (Choi & Darwiche, 2006a). First, note that $F_\beta = U_\beta - H_\beta$ where $U_\beta$ is the Bethe *average* energy

$$U_\beta = -\sum_{(i,j) \in \mathcal{E}} E[\log \psi(X_i, X_j)] - \sum_{i \in \mathcal{V}} E[\log \psi(X_i)]$$

and where $H_\beta$ is the Bethe *approximate* entropy

$$H_\beta = \sum_{(i,j) \in \mathcal{E}} H(X_i, X_j) - \sum_{i \in \mathcal{V}} (n_i - 1) H(X_i)$$

where $n_i$ is the number of neighbors of node $i$ in $\mathcal{M}$ (for details, see Yedidia et al., 2005).

It will be convenient to start with the case where every edge $(i,j)$ in the unextended model is replaced with a chain $\{(i,i'),(i',j'),(j',j)\}$. We then delete all equivalence edges $(i,i'), (j',j) \in \mathcal{E}^\star$. Note that the resulting network $\mathcal{M}'$ has $n + 2m$ nodes: $n$ nodes $i \in \mathcal{V}$, and 2 clone nodes $i', j'$ for each of the $m$ edges $(i',j') \in \mathcal{E}$.

The average energy $U'$ and the entropy $H'$ for $\mathcal{M}'$ is

$$U' = -\sum_{(i',j') \in \mathcal{E}} E[\log \psi(X_i, X_j)] - \sum_{i \in \mathcal{V}} E[\log \psi(X_i)]$$
$$- \sum_{(i,i') \in \mathcal{E}^\star} E[\log \theta(X_i)\theta(X_i')]$$
$$H' = \sum_{(i',j') \in \mathcal{E}^\star} H(X_i, X_j) + \sum_{i \in \mathcal{V}} H(X_i).$$

Since $\theta(x_i)\theta(x_j) = z_{ij} Pr(x_i)$ (see Equation 2), we have

$$E[\log \theta(X_i)\theta(X_i')] = \log z_{ij} - H(X_i). \quad (5)$$

We can show through further manipulations that

$$\sum_{(i,i') \in \mathcal{E}^\star} E[\log \theta(X_i)\theta(X_i')] = \log z - \sum_{i \in \mathcal{V}} n_i H(X_i).$$

After substituting into $U'_\beta$, and some rearrangement:

$$F' = U' - H' = U_\beta - H_\beta - \log z = F_\beta - \log z$$

as desired. To show this correspondence continues to hold for any tree-structured $\mathcal{M}'$, we note first that IBP beliefs continue to be node and edge marginals for any tree-structured ED-BP approximation $\mathcal{M}'$. Next, when we recover an edge into a tree approximation that yields another tree approximation, we lose an expectation over edge parameters (Equation 5). The corresponding node entropy $H(X_i)$ that is lost in the average energy $U'$ is canceled out by a node entropy gained in the entropy $H'$. Finally, the term $\log z_{ij}$ that is lost is no longer needed in the correction factor $z$ after recovery. Thus, we can recover edges into our fully disconnected approximation, and conclude that $F' = F_\beta - \log z$ continues to hold for any tree-structured approximation $\mathcal{M}'$. $\square$

**Proof of Proposition 2** In an extended network $\mathcal{M}$ with equivalence edge $(i,j)$ and potential $\phi(x_i, x_j)$:

$$Z = \sum_{x_i = x_j} \frac{\partial Z}{\partial \phi(x_i, x_j)} = \sum_{x_i = x_j} \frac{\partial^2 Z'}{\partial \theta(x_i)\partial \theta(x_j)}$$
$$= \sum_{x_i = x_j} \frac{Z' Pr'(x_i, x_j)}{\theta(x_i)\theta(x_j)} = \sum_{x_i = x_j} \frac{Z' Pr'(x_i, x_j)}{z_{ij} Pr'(x_j)}$$
$$= \frac{Z'}{z_{ij}} \sum_{x_i = x_j} Pr'(x_i \mid x_j)$$

which is simply $Z' \cdot \frac{y_{ij}}{z_{ij}}$. Note that the fourth equality follows from Equation 2. $\square$

**Proof of Proposition 3** In an extended network $\mathcal{M}$ with equivalence edges $(i,j)$ and $(s,t)$ and edge potentials $\phi(x_i, x_j)$ and $\phi(x_s, x_t)$:

$$Z = \sum_{\substack{x_i = x_j \\ x_s = x_t}} \frac{\partial^2 Z}{\partial \phi(x_i, x_j) \partial \phi(x_s, x_t)}$$

$$= \sum_{\substack{x_i = x_j \\ x_s = x_t}} \frac{\partial^4 Z'}{\partial \theta(x_i) \partial \theta(x_j) \partial \theta(x_s) \partial \theta(x_t)}$$

$$= \sum_{\substack{x_i = x_j \\ x_s = x_t}} \frac{Z' Pr'(x_i, x_j, x_s, x_t)}{\theta(x_i) \theta(x_j) \theta(x_s) \theta(x_t)}$$

$$= \sum_{\substack{x_i = x_j \\ x_s = x_t}} \frac{Z' Pr'(x_i, x_j, x_s, x_t)}{z_{ij} Pr'(x_j) z_{st} Pr'(x_t)} \quad \text{by Eq. 2}$$

$$= \frac{Z'}{z_{ij} z_{st}} \sum_{\substack{x_i = x_j \\ x_s = x_t}} \frac{Pr'(x_i, x_j) Pr'(x_s, x_t)}{Pr'(x_j) Pr'(x_t)}$$

$$= \frac{Z'}{z_{ij} z_{st}} \sum_{x_i = x_j} Pr'(x_i | x_j) \sum_{x_s = x_t} Pr'(x_s | x_t)$$

which is simply $Z' \cdot \frac{y_{ij}}{z_{ij}} \frac{y_{st}}{z_{st}}$. □